\documentclass[runningheads]{article}
\usepackage[utf8]{inputenc}
\usepackage{hyperref}
\usepackage{times}
\usepackage[numbers, sort&compress, sectionbib]{natbib}
\usepackage{latexsym}
\usepackage{amsmath}
\usepackage{tabularx,ragged2e}
\usepackage{booktabs}
\usepackage{url}
\usepackage{graphicx}
\usepackage{todonotes}
\usepackage{subcaption}
\usepackage{caption}
\usepackage{float}
\usepackage[english]{babel}
\usepackage{multirow}

\newcolumntype{L}{>{\RaggedRight\arraybackslash}X}

\newcommand*{\centerfloat}{%
  \parindent \z@
  \leftskip \z@ \@plus 1fil \@minus \textwidth
  \rightskip\leftskip
  \parfillskip \z@skip}
\makeatother

\title{Time to Take Emoji Seriously: They Vastly Improve Casual Conversational Models}

\author{Pieter Delobelle and Bettina Berendt\\
  {\small KU Leuven, Department of Computer Science}\\
  {\small Celestijnenlaan 200A box 2402, 3000 Leuven, Belgium}\\
  {\small \texttt{\{pieter.delobelle, bettina.berendt\}@cs.kuleuven.be}}}
\date{}

\begin{document}
\maketitle
\begin{abstract}
  Graphical emoji are ubiquitous in modern-day online conversations. So is a single thumbs-up emoji able to signify an agreement, without any words. We argue that the current state-of-the-art systems are ill-equipped to correctly interpret these emoji, especially in a conversational context. However, in a casual context, the benefits might be high: a better understanding of users' utterances and more natural, emoji-rich responses. 

  With this in mind, we modify BERT to fully support emoji, both from the Unicode Standard and custom emoji. This modified BERT is then trained on a corpus of question-answer (QA) tuples with a high number of emoji, where we're able to increase the \emph{1-of-100 accuracy} from 12.7\% for the current state-of-the-art to 17.8\% for our model with emoji support. 
\end{abstract}

\let\thefootnote\relax\footnotetext{Copyright © 2019 for this paper by its authors. Use permitted under Creative Commons License Attribution 4.0 International (CC BY 4.0).}

\section{Introduction}\label{sec:intro}
The prevalent use of emoji---and their text-based precursors---is mostly unaddressed in current natural language processing (NLP) tasks. The support of the Unicode Standard~\citep{theunicodeconsortiumUnicodeStandard2010} for emoji characters in 2010 ushered in a wide-spread, international adoption of these graphical elements in casual contexts. Interpreting the meaning of these characters has been challenging however, since they take on multiple semantic roles~\citep{naamanVaryingLinguisticPurposes2017}.

Whether or not emoji are used depends on the context of a text or conversation, with more formal settings generally being less tolerating. So is the popular aligned corpus Europarl~\citep{koehnEuroparlParallelCorpus2012} naturally devoid of emoji. Technical limitations, like no Unicode support, also limit its use. This in turn affects commonly used corpora, tokenizers, and pre-trained networks.

Take for example the Ubuntu Dialog Corpus by \citet{loweUbuntuDialogueCorpus2015}, a commonly used corpus for multi-turn systems. This dataset was collected from an Internet Relay Chat (IRC) room casually discussing the operating system Ubuntu. IRC nodes usually support the ASCII text encoding, so there's no support for graphical emoji. However, in the 7,189,051 utterances, there are only 9946 happy emoticons (i.e. {\tt:-)} and the cruelly denosed {\tt:)} version) and 2125 sad emoticons.

Word embeddings are also handling emoji poorly: Word2vec~\citep{mikolov2013} with the commonly used pre-trained Google News vectors doesn't support the graphical emoji at all and vectors for textual emoticons are inconsistent. As another example with contextualized word embeddings, there are also no emoji or textual emoticons in the vocabulary list of BERT~\citep{devlin2019} by default and support for emoji is only recently added to the tokenizer. The same is true for GPT-2~\citep{radford2019language}. As all downstream systems, ranging from multilingual r\'esum\'e parsing to fallacy detection~\citep{delobelle2019}, rely on the completeness of these embeddings, this lack of emoji support can affect the performance of some of these systems.

Another challenge is that emoji usage isn't static. Think of shifting conventions, different cultures, and newly added emoji to the Unicode list. Several applications also use their own custom emoji, like chat application Slack and streaming service Twitch. This becomes an issue for methods that leverage the Unicode description~\citep{eisnerEmoji2vecLearningEmoji2016} or that rely on manual annotations~\citep{tigwellOhThatWhat2016}. 

Our contribution with this paper is two-fold: firstly, we argue that the current use---or rather non-existing use---of emoji in the tokenizing, training, and the datasets themselves is insufficient. Secondly, we attempt to quantify the significance of incorporating emoji-based features by presenting a fine-tuned model. We then compare this model to a baseline, but without special attention to emoji.

Section~\ref{sec:related-work} will start with an overview of work on emoji representations, emoji-based models and analysis of emoji usage. A brief introduction in conversational systems will also be given. Section~\ref{sec:data} will then look into popular datasets with and without emoji and then introduce the dataset we used. 

Our model will then be discussed in Section~\ref{sec:experiment}, including the tokenization in Subsection~\ref{ss:tokenization}, training setup in Subsection~\ref{ss:training} and evaluation in Subsection~\ref{ss:evaluation}. This brings us to the results of our experiment, which is discussed in Section~\ref{sec:results} and finally our conclusion and future work are presented in Section~\ref{sec:conclusion}.

\section{Related work}\label{sec:related-work}
Inspired by the work on word representations, \citet{eisnerEmoji2vecLearningEmoji2016} presented Emoji2vec. This system generates a vector representation that's even compatible with the Word2vec representations, so they can be used together. This compatibility makes it easy to quickly incorporate Emoji2vec in existing systems that use Word2vec. 

The main drawback is that the emoji representations are trained on the Unicode descriptions. As a consequence, the representations only capture a limited meaning and do not account for shifting or incorrect use of emoji in the real world. For example, a peach emoji could be considered a double entendre, due to the resemblance to a woman's posterior. This is of course mentioned nowhere in the Unicode description. Which shows that the meaning of an emoji is how users interpret it, so also accidental incorrect use can cause issues~\citep{millerUnderstandingEmojiAmbiguity2017}.

In spirit, \citet{felboUsingMillionsEmoji2017} is similar to our work. Their system, DeepMoji, illustrates the importance of emoji for sentiment, emotion, and sarcasm classification. For these tasks, they used a dataset of 1246 million tweets containing at least one emoji. However, the authors use the emoji in those tweets not for the DeepMoji model input, but as an target label. With a slightly better agreement score than humans on the sentiment task, this supports our hypothesis that emoji carry the overall meaning of an utterance. 

\citet{barbieriWhatDoesThis2016} focus on a predicting one emoji based on the textual content. Interestingly, they looked into both English and Spanish tweets and compared a range of systems for a shared task at SemEval 2018: Multilingual Emoji Prediction. This shared task shows that emoji are getting more attention, but how their task is set up also highlights the current lack of high quality datasets with emoji. 

The same shared task was tackled by~\citep{chenPeperomiaSemEval2018Task2018} and a year later by~\citet{huangANASemEval2019Task2019}, which made use of a pre-processor and tokenizer from~\citet{baziotisDataStoriesSemEval2017Task2017}. This tokenizer replaces some emoji and emoticons by tokens related to their meaning. So is {\tt \char`\\o/} replaced with {\tt <happy>}. Naturally, this approach suffers from the same issues as described before. And even though it's really useful to have some basic, out-of-the-box support for emoticons thanks to this work, we think that this strategy is too reducing to capture subtle nuances.

An analysis on the use of emoji on a global scale is done by~\citet{ljubesicGlobalAnalysisEmoji2016}. For this, the authors used geo-tagged tweets, which also allowed them to correlate the popularity of certain emoji with development indicators. This shows that the information encoded by emoji---and of course the accompanying tweet---is not limited to sentiment or emotion. Also \citet{ai2017untangling} analyze the uses of emoji on social networks. Their approach consists of finding information networks between emoji and English words with LINE~\citep{tang2015line}.

An interesting aspect of emoji usage is analyzed by \citet{robertson2018self}. In this work, the correlation between the use of Fitzpatrick skin tone~\citep{fitzpatrickValidityPracticalitySunReactive1988} modifiers and the perceived skin tone of the user. This research shows that users are inclined to use representing emoji for themselves. \citet{robertson2018self} reported that no negative sentiment was associated with specific skin tone modifiers.

\begin{table*}[tb]
\centering
\caption{Overview of the support for both graphical emoji and textual emoticons for common corpora. No support is indicated with an em bar (---). Otherwise, we provide the fraction of utterances with at least one emoji or emoticon.}\label{tab:datasets}
\begin{tabular}{@{}lrrrr@{}}
\toprule
Dataset                                                           & Size                  & ~~~Language     & ~~~Emoji        & ~~~Emoticons \\ \midrule
Ubuntu~\citep{loweUbuntuDialogueCorpus2015}                       & 7,189,051             & EN              & ---             & 0.17\%            \\
Amazon QA~\citep{wanModelingAmbiguitySubjectivity2016}            & 1,396,896             & EN              & ---             & 0.14\%            \\
OpenSubtitles~\citep{lisonOpenSubtitles2016ExtractingLarge2016}   & 316,891,717           & EN              & ---             & ---               \\
ConvAI2 Persona-Chat~\citep{zhangPersonalizingDialogueAgents2018} & 47,234                & EN              & ---             & 1.25\%            \\ \midrule 
Twitter customer support~\citep{horiEndtoendConversationModeling2017} & 2,000                 & EN              & 8.75\%          & 7.50\%            \\ \bottomrule
\end{tabular}
\end{table*}

\paragraph{Conversational AI systems}
The research on conversational AI has been focussing on various aspects, including building high-quality datasets~\citep{loweUbuntuDialogueCorpus2015, kummerfeldLargeScaleCorpusConversation2019, lisonOpenSubtitles2016ExtractingLarge2016, zhangPersonalizingDialogueAgents2018, hendersonRepositoryConversationalDatasets2019, reddyCoQAConversationalQuestion2019}, adding customizable personalities~\citep{zhangPersonalizingDialogueAgents2018, zhangModelingMultiturnConversation2018, mazareTrainingMillionsPersonalized2018} or conjoining the efforts with regard to different datasets, models and evaluation practices~\citep{hendersonRepositoryConversationalDatasets2019}. With these combined efforts, businesses and the general public quickly began developing ambitious use-cases, like customer support agents on social networks.

The proposed models in this field are diverse and largely depending on how the problem is formulated. When considering free-form responses, generative models like GPT~\citep{radfordImprovingLanguageUnderstanding2018}, GPT-2~\citep{radford2019language} or seq2seq~\citep{sutskeverSequenceSequenceLearning2014a} are appropriate. When the conversational task is modeled as a response selection task to pick the correct response out of $N$ candidates~\citep{hendersonEfficientNaturalLanguage2017, hendersonRepositoryConversationalDatasets2019, yu-etal-2016-strategy}, this can be a language model like BERT~\citep{devlin2019} with a dedicated head.


\section{Emoji-rich datasets are hard to find}\label{sec:data}

Emoji are commonly used in casual settings, like on social media or in casual conversations. In conversations---as opposed to relatively context-free social media posts---an emoji alone can be an utterance by itself. And with a direct impact for some applications, like customer support, we focus on conversational datasets. We hope the conversational community has the most direct benefit from these emoji-enabled models. Of course, the conclusions we'll draw don't have to be limited to this field.

Table~\ref{tab:datasets} gives an overview of frequently used and interesting conversational datasets. The lacuna of emoji-rich reference datasets was already mentioned in Section~\ref{sec:intro} and is in our opinion one of the factors that emoji remain fairly underutilized. 

For our models, we'll use a customer support dataset with a relatively high usage of emoji. The dataset contains 2000 tuples collected by~\citet{horiEndtoendConversationModeling2017} that are sourced from Twitter. They provide conversations, which consist of at least one question and one free-form answer. Some conversations are longer, in this case we ignored the previous context and only looked at the last tuple. This dataset illustrates that even when contacting companies, Twitter users keep using emoji relatively often, 8.75\% of all utterances. 

The tweets were filtered on hyper links and personal identifiers, but Unicode emoji characters were preserved. As emoji are frequently used on Twitter, this resulted in a dataset with 170 of the 2000 tuples containing at least one emoji character. 


\begin{figure}[tb]
  \centering
  \includegraphics[width=1.1\textwidth]{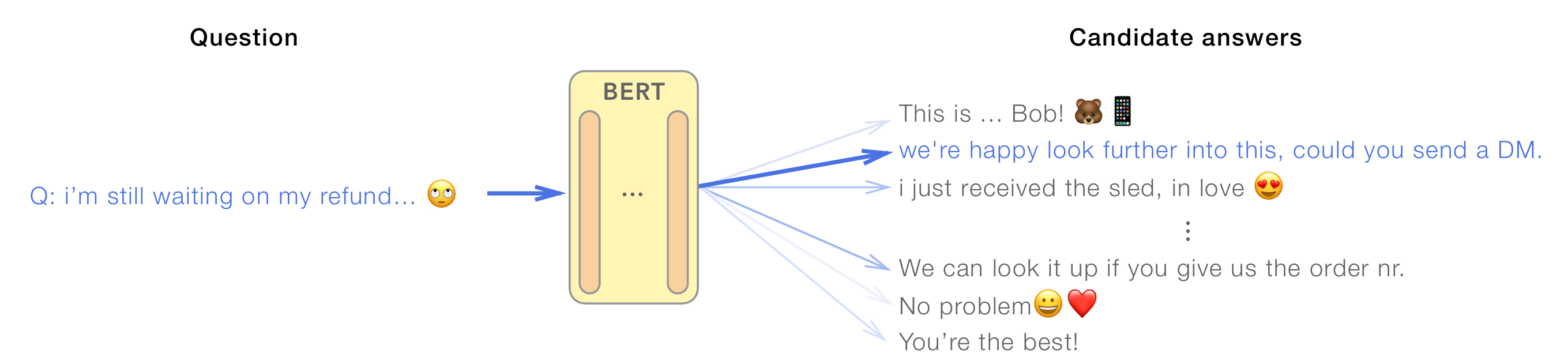}
\caption{Illustration of the use of BERT for a response selection task.}
\label{fig:bert-resp}
\end{figure}

\section{Fine-tuning BERT with emoji support}\label{sec:experiment}
We continue training of a multilingual BERT model~\citep{devlin2019} with new tokens for emoji and fine-tune this model and a baseline on the dataset discussed in Section~\ref{sec:experiment}. This approach is explained in Subsection~\ref{ss:tokenization} and the training itself is discussed in Subsection~\ref{ss:training}. At last, the evaluation is then discussed in Subsection~\ref{ss:evaluation}.

\subsection{Tokenizing emoji}\label{ss:tokenization}
We add new tokens to the BERT tokenizer for 2740 emoji from the Unicode Full Emoji List~\citep[v12.0]{theunicodeconsortiumUnicodeStandard2010}, as well as some aliases (in the form of {\tt :happy:} as is a common notation for emoji). In total, 3627 emoji tokens are added to the vocabulary. 

We converted all UTF-8 encoded emoji to a textual alias for two reasons. First, this mitigates potential issues with text encodings that could drop the emoji. Second, this is also a common notation format for custom emoji, so we have one uniform token format. Aside from this attention to emoji, we use WordPiece embeddings~\citep{wuGoogleNeuralMachine2016} in the same manner as \citet{devlin2019}.

\begin{figure}[tb]
  \centering
  \includegraphics[width=1.1\textwidth]{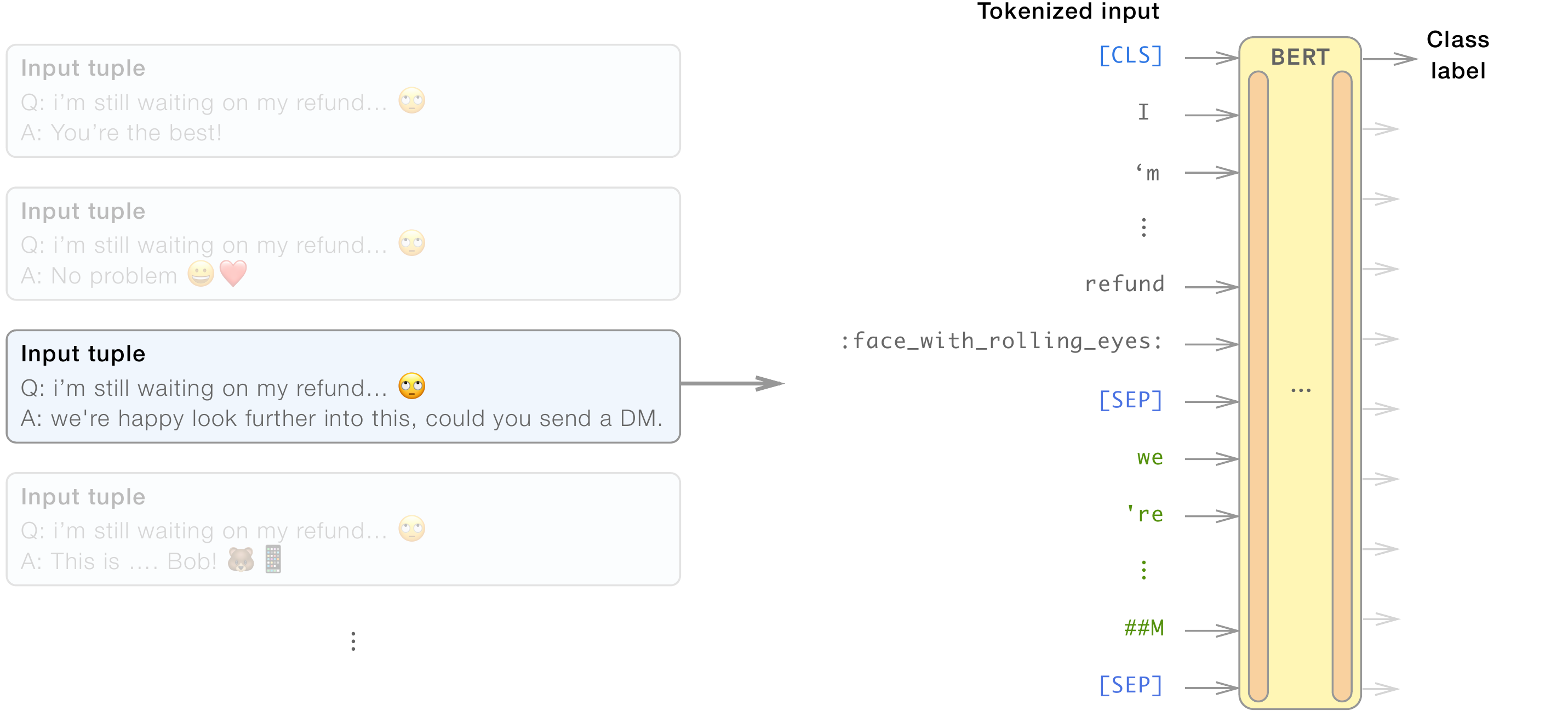}
\caption{Illustration the  tokenization process for BERT. The illustrated input is an actual sample from the test set. Notice the emoji is replaced by a descriptive token.}
\label{fig:bert-token}
\end{figure}

\subsection{Training and fine-tuning}\label{ss:training}
We start from 12-headed multilingual BERT ({\tt bert-base-multilingual-cased}), which has 110M parameters. For the model with emoji support, the number of tokens is increased, so new vectors are appended at the end of the embeddings matrix. We then continue training on the language modeling task. We use the default configuration as is also used by \citet{devlin2019} where randomly selected tokens are replaced by:

\begin{itemize}
  \item \textbf{a mask token}: 80\% chance,
  \item \textbf{another random word}: 10\% chance,
  \item \textbf{the original word}: 10\% chance.
\end{itemize}

This model is trained for 100 epochs with the Adam~\citep{kingmaAdamMethodStochastic2014} optimizer. The learning rate is set to the commonly used $lr=5\cdot10^{-5}$ and $\epsilon = 10^{-8}$. No hyper-parameter tuning was done, as the results are acceptable on their own and are sufficient to allow conclusions for this paper. The loss is cross entropy~\cite{janochaLossFunctionsDeep2017}.

We then fine-tune both models, with and without emoji tokenization, on the sentence prediction task with a training set of 70\%. We again use the Adam optimizer with the same settings and with binary cross entropy. In this case, the training was limited to 10 epochs. To mitigate the need for weighting and other class imbalance issues, we trained with pairs of positive and negative candidates. This is in contract to the evaluation, where 99 negative candidates are used. However, since each candidate is considered on its own merit during evaluation, this discrepancy won't affect the performance.

For the formulation of the fine-tuning task, we use the same approach as \citet{devlin2019}. The first input sentence is joined with the second sentence, separated by a special {\tt [SEP]} token, as can be seen in Figure~\ref{fig:bert-token}. The model, with a specialized head for \emph{next sentence prediction}, then outputs a correlation score.

\subsection{evaluation metrics}\label{ss:evaluation}
Finally, our model is compared against the pre-trained version of BERT without special emoji tokens. We evaluate both this baseline and our model as a response selection task. In this case, the system has to select the most appropriate response out $N=100$ candidates. This is a more restricted problem, where the \emph{1-of-100 accuracy}~\citep{hendersonRepositoryConversationalDatasets2019} is a popular evaluation metric.

Note that 1-in-100 accuracy gives a summary of the model performance for a particular dataset. Since not all 99 negative responses are necessarily bad choices, the resulting score is in part dependent on the prior distribution of a dataset. For example, \citet{hendersonRepositoryConversationalDatasets2019} compares models for three datasets, where the best performing model has a score of 30.6 for OpenSubtitles~\cite{lisonOpenSubtitles2016ExtractingLarge2016} and 84.2 for AmazonQA~\cite{wanModelingAmbiguitySubjectivity2016}. 

Aside from the 1-of-100 accuracy, we also present the mean rank of the correct response. Since the Twitter dataset is focussed on customer service, the correct response is sometimes similar to others. The mean rank, also out of $N=100$, can differentiate whether or not the model is still selecting good responses. For each input sentence, a rank of 1 means the positive response is ranked highest and is thus correctly selected and a rank of $N$ signifies the positive response was---incorrectly---considered the worst-matching candidate.

\section{Emoji provide additional context to response selection models}\label{sec:results}
After training of the language model with additional tokens for all Unicode emoji, we achieved a final perplexity of 2.0281. For comparison, the BERT model with 16 heads achieved a perplexity of 3.23~\citep{devlin2019}, but this is on a general dataset.

For the sentence prediction task, Table~\ref{tab:results} shows the results of the baseline and our model with additional emoji tokens. For each of the 600 utterance pairs of the held-out test set, we added 99 randomly selected negative candidates, as described in Subsection~\ref{ss:evaluation}. The \emph{1-out-of-100 accuracy} measures how often the true candidate was correctly selected and the \emph{mean rank} gives an indication of how the model performs if it fails to correctly select the positive candidate.

\begin{table}[tb]
\centering
\caption{1-of-100 accuracy and mean ranking out of 100 candidates on the test set of 600 utterance tuples.}
\label{tab:results}
\resizebox{\textwidth}{!}{%
\begin{tabular}{@{}lrr@{}}
\toprule
Model                                                & 1-of-100 ACC    & Mean rank \\ \midrule
{\tt bert-base-multilingual-cased}                   & 12.7\%          & 33        \\
{\tt bert-base-multilingual-cased} with emoji tokens & \textbf{17.8\%} & \textbf{31}\\ \bottomrule
\end{tabular}%
}
\end{table}

The baseline correctly picks 12.7\% of all candidate responses, out of 100. Given that the dataset is focussed on support questions and multiple responses are likely to be relevant, this baseline already performs admirable. For reference, a BERT model on the OpenSubtitles dataset~\citep{lisonOpenSubtitles2016ExtractingLarge2016} achieves a \emph{1-of-100 accuracy} between 12.2\% and 17.5\%, depending on the model size~\citep{hendersonRepositoryConversationalDatasets2019}.

Our model improves on this baseline with a \emph{1-of-100 accuracy} of 17.8\%. The mean rank remains almost the same. This indicates that the emoji tokens do help with with picking the correct response, but don't really aide when selecting alternative suitable candidates. One possible explanation is that when emoji are used (this is the case for 8.75\% of all utterances), including those tokens helps matching those based on those emoji and their meaning. When there are no emoji present, our model might be just as clueless as the baseline.  


\section{Conclusion and future work}\label{sec:conclusion}
In this paper we discussed the current state of emoji usage for conversational systems, which mainly lacks large baseline datasets. When looking at public datasets, conversational AI makers have to choose between dataset size and emoji support, with some datasets at least containing a few textual emoticons. We argued that this duality results in systems that fail to capture some information encoded in those emoji and in turn fail to respond adequately. 

Based on this premise, we investigated how a response selection system based on BERT can be modified to support emoji. We proposed a format and tokenization method that's indifferent to current Unicode specifications, and thus also works for datasets containing custom emoji.

Evaluation of this emoji-aware system increased the \emph{1-of-100 accuracy} from 12.7\% for the baseline to 17.8\%. Thus showing that supporting emoji correctly can help increasing performance for more casual systems, without having to rely on labeling or external descriptions for those emoji.

However, the lack of high-quality, general datasets with emoji limits our conversational model. Working towards larger casual conversational datasets would help both for our model, and for the conversational NLP community in general.

We investigated the impact of emoji for conversational models and one could argue that these conclusions---or even the BERT model---can be generalized. We didn't investigate whether other tasks also benefited from our fine-tuned BERT model with the additional emoji tokens.

During evaluation, we also observed utterances with only emoji characters. Even with our model that supports emoji, it could still be difficult to extract information like the subject of a conversation. Some of these utterances---but not all---were part of a larger conversation, so an interesting question could be how additional information affects the model.  

\vspace{-1em}
\section*{Acknowledgements}

\vspace{-0.5em}
This work was supported by the Research Foundation - Flanders under EOS No. 30992574.

\vspace{-0.5em}

\bibliographystyle{plainnat}
\bibliography{bibliography}
\end{document}